\documentclass[letterpaper, 10 pt, conference]{ieeeconf}  

\IEEEoverridecommandlockouts                              

\usepackage{graphicx, tipa}
\usepackage{amsmath,amssymb,amsfonts}
\usepackage[caption=false]{subfig}
\usepackage[ruled,linesnumbered, noend]{algorithm2e}
\usepackage{dirtytalk}
\usepackage{hyperref}
\usepackage{gensymb}
\usepackage{tabularx}
\usepackage{multirow}
\usepackage[short]{optidef}
\usepackage{booktabs}
\usepackage{siunitx}
\usepackage{titlesec}
\usepackage{amssymb} 

\newcounter{cases}
\newcounter{subcases}[cases]

\makeatletter
\newcommand{\removelatexerror}{\let\@latex@error\@gobble}
\makeatother
\SetKwComment{Comment}{$\triangleright$\ }{}

\newcommand\Tstrut{\rule{0pt}{2.0ex}}         

\title{\huge
LV-DOT: LiDAR-visual dynamic obstacle detection and tracking for autonomous robot navigation}

\author{Zhefan Xu\footnotemark*, Haoyu Shen\footnotemark*, Xinming Han, Hanyu Jin, Kanlong Ye, and Kenji Shimada 
\thanks{*The authors contributed equally.
    \newline{\indent Zhefan Xu, Haoyu Shen, Xinming Han, Hanyu Jin, Kanlong Ye, and Kenji Shimada are with the Department of Mechanical Engineering, Carnegie Mellon University, 5000 Forbes Ave, Pittsburgh, PA, 15213, USA.}
        {\tt\small zhefanx@andrew.cmu.edu}}%
}

\begin{document}

\maketitle
\thispagestyle{empty}
\pagestyle{empty}

\begin{abstract}
Accurate perception of dynamic obstacles is essential for autonomous robot navigation in indoor environments. Although sophisticated 3D object detection and tracking methods have been investigated and developed thoroughly in the fields of computer vision and autonomous driving, their demands on expensive and high-accuracy sensor setups and substantial computational resources from large neural networks make them unsuitable for indoor robotics. Recently, more lightweight perception algorithms leveraging onboard cameras or LiDAR sensors have emerged as promising alternatives. However, relying on a single sensor poses significant limitations: cameras have limited fields of view and can suffer from high noise, whereas LiDAR sensors operate at lower frequencies and lack the richness of visual features. To address this limitation, we propose a dynamic obstacle detection and tracking framework that uses both onboard camera and LiDAR data to enable lightweight and accurate perception. Our proposed method expands on our previous ensemble detection approach, which integrates outputs from multiple low-accuracy but computationally efficient detectors to ensure real-time performance on the onboard computer. In this work, we propose a more robust fusion strategy that integrates both LiDAR and visual data to enhance detection accuracy further. We then utilize a tracking module that adopts feature-based object association and the Kalman filter to track and estimate detected obstacles' states. Besides, a dynamic obstacle classification algorithm is designed to robustly identify moving objects. The dataset evaluation demonstrates a better perception performance compared to benchmark methods. The physical experiments\footnote{Experiment video link: \url{https://youtu.be/rRvgTulWqvk}} on a quadcopter robot confirms the feasibility for real-world navigation.
\end{abstract}

\section{Introduction}
Deploying robots in indoor workspaces requires the accurate perception of dynamic obstacles. However, the limited sensor capabilities and computational resources of mobile robots make real-time dynamic obstacle perception challenging. Developing a robust and efficient dynamic obstacle perception algorithm is therefore essential for robot navigation.

The challenges of dynamic obstacle perception for mobile robots can be summarized into three key aspects. First, unlike autonomous vehicles that benefit from powerful GPUs and high-accuracy sensor combinations, mobile robots are constrained by limited computational resources and sensor quality, making computationally intensive learning-based methods \cite{second}\cite{pv-rcnn} impractical. Second, detecting obstacles from onboard sensor data in cluttered environments often results in significant inaccuracies. Although previous work has adopted various approaches such as image-based \cite{vision_ccmpc}, point clustering-based \cite{zju_dynamic_avoidance}, and map-based detection methods \cite{dsp} to improve detection accuracy, false detections remain prevalent. Finally, identifying dynamic obstacles among detected ones is challenging due to sensor noises. Existing identification methods that incorporate occlusion principles \cite{m-detector} and occupancy updates \cite{dsp}\cite{dynablox} can sometimes be overly conservative, leading to misdetections and consequently high-latency responses.

\begin{figure}[t] 
    \vspace{0.25cm}
    \centering
    \includegraphics[scale=0.425]{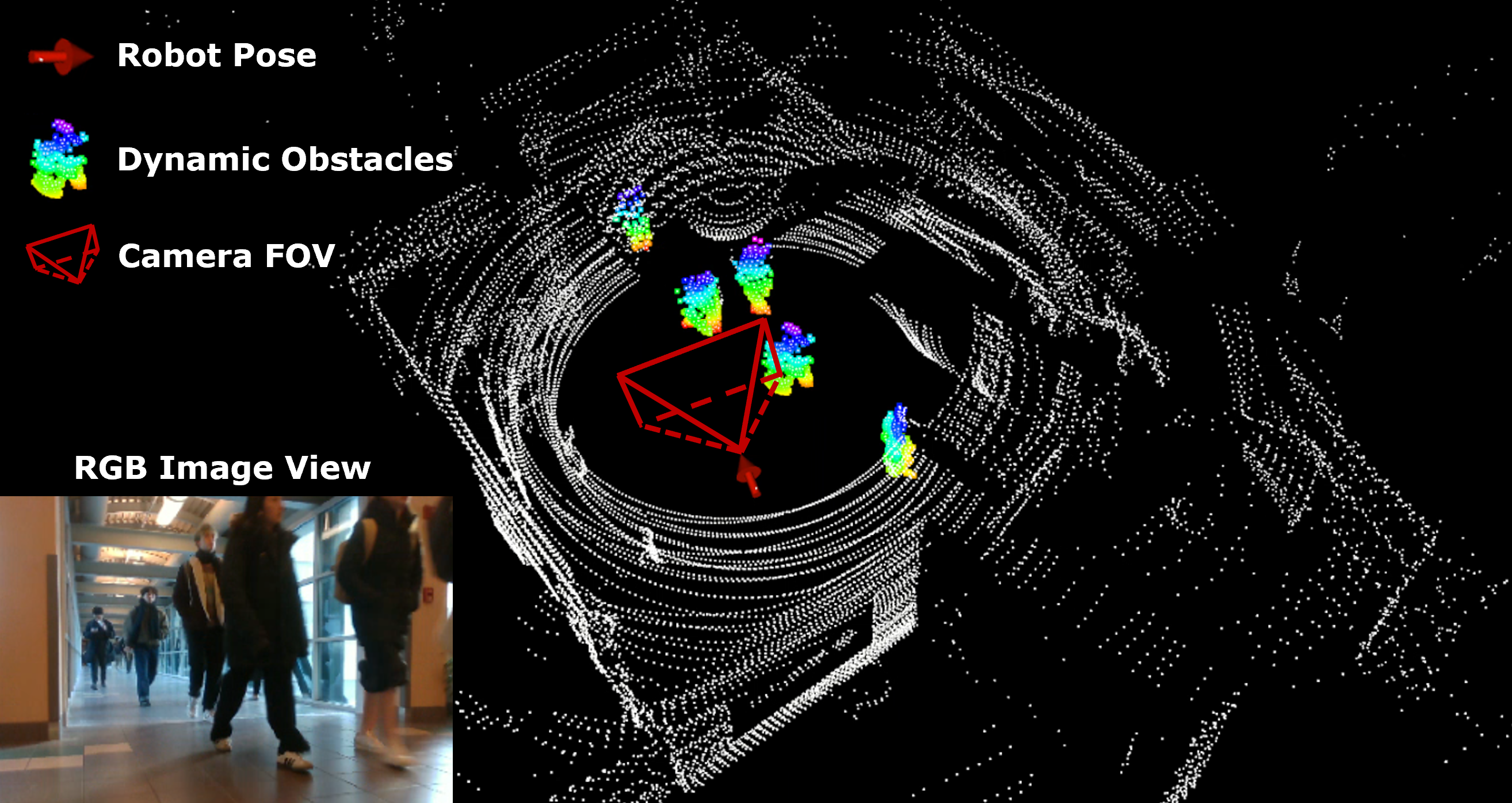}
    \caption{Visualization of dynamic obstacle perception using the proposed framework, with the point cloud of tracked dynamic obstacles highlighted. }
    \label{intro_figure}
\end{figure}

To mitigate these issues, we propose a lightweight dynamic obstacle detection and tracking framework that leverages both an onboard RGB-D camera and LiDAR. The underlying idea is that LiDAR provides a longer and wider (360-degree) perception range with high depth measurement accuracy, while the camera delivers rich visual features that improve dynamic object identification and supplies higher frequency and denser depth information in close proximity. To achieve this, we first adopt the proposed LiDAR and camera obstacle detection algorithm to obtain unclassified obstacle clusters with bounding boxes from the LiDAR point cloud and RGB-D image, respectively. Next, we extend our ensemble detection method \cite{visual_detection} to fuse the detection results from both sensors, thereby achieving higher detection accuracy. Finally, a tracking and dynamic object identification module is applied to classify the detected obstacles into static and dynamic categories. Fig. \ref{intro_figure} shows an example of dynamic obstacle detection and tracking using the proposed method. The main contributions of this work are as follows: 

\begin{itemize}
    \item \textbf{The LV-DOT framework:} We introduce a novel LiDAR-visual dynamic obstacle detection and tracking (LV-DOT) framework that leverages the complementary strengths of LiDAR and RGB-D camera. LiDAR enhances the detection range and accuracy of the camera, while the camera contributes rich visual features and higher-frequency data to complement the LiDAR. The LV-DOT perception framework is made available as an open-source ROS package on GitHub \footnote{\url{https://github.com/Zhefan-Xu/LV-DOT}}.
    \item \textbf{High computational efficiency:} Our approach adopts an ensemble detection strategy that achieves high computational efficiency by integrating several obstacle detectors that have low accuracy but high efficiency.
    \item \textbf{Physical robot experiments:} Experiments on physical robots conducted in various environments demonstrate that the proposed framework can be effectively deployed to ensure safe navigation in dynamic environments.
\end{itemize}

\section{Related Work}
In recent years, extensive efforts have been made to develop dynamic obstacle perception algorithms using LiDAR and camera sensors. Although numerous learning-based methods \cite{second}\cite{pv-rcnn}\cite{dot}\cite{ea-lss} have been proposed in the autonomous driving domain, their relatively high computational requirements make them less suitable for mobile robots. Consequently, this section mainly focuses on lightweight methods suitable for mobile robots, and we categorize these approaches into camera-based and LiDAR-based methods.

\textbf{Camera-based methods:} Methods in this category utilize color and depth images to perceive obstacles. In \cite{eth_yolo_detection}, a clustering algorithm is applied to points projected from the depth image, while a 2D people detector identifies dynamic obstacles. Following a similar clustering approach, Wang et al. \cite{zju_dynamic_avoidance} propose a dynamic classification method for generic obstacles based on point distances in consecutive frames. Lin et al. \cite{vision_ccmpc} adopt the U-depth detector \cite{first_udepth} to efficiently detect unclassified obstacles and demonstrate safe collision avoidance. This concept is further extended in \cite{dynamic_map_ours}\cite{saha2022efficient} by incorporating a tracking module that classifies obstacles based on estimated velocities. In \cite{lu2022perception}, Lu et al. effectively detect small dynamic obstacles by applying the 3D SORT algorithm, while Chen et al. \cite{chen2022real} enhance tracking performance under occlusion using obstacle feature vectors and track points. To address false positives arising from noisy data, the ensemble detection method \cite{visual_detection} improves dynamic obstacle detection accuracy by finding mutual agreement among lightweight detectors. Moreover, in \cite{dsp}, the occupancy map is extended to allow fast occupancy updates for dynamic obstacles. Beyond obstacle avoidance, dynamic obstacle detection is applied in SLAM to improve robustness, with dynamic obstacle removal methods adopted in \cite{sun2017improving}\cite{shi2018dynamic}\cite{ds_slam}\cite{dai2020rgb} to enhance RGB-D SLAM accuracy in dynamic environments.

\textbf{LiDAR-based methods:} LiDAR-based methods typically use point clouds for obstacle perception. Early techniques \cite{asvadi20163d}\cite{yoon2019mapless} adopt consecutive frame point cloud subtraction to identify dynamic points, followed by post-processing to refine the results. In \cite{dewan2016motion}, a motion cue combined with a Bayesian approach is introduced to detect dynamic points from LiDAR scans. Min et al. \cite{min2021kernel} propose a kernel-based dynamic map to differentiate dynamic obstacles in the occupancy map, while Zhou et al. \cite{zhou2023lidar} filter dynamic obstacles from the navigation map based on the assumption that they are in contact with and located on the ground. Recently, Schmid et al. \cite{dynablox} extended the voxel map approach to identify dynamic obstacles using voxel status, demonstrating success in indoor environments. Additionally, Wu et al. \cite{m-detector} utilizes the principle of occlusion to classify points from LiDAR scan as dynamic, and Saha et al. \cite{saha20243d} convert LiDAR scans into multiple depth map representations, enabling image-based detection with dynamic obstacle classification through tracking and velocity estimation. In \cite{du2024lidar}, Du et al. applies a point cloud clustering algorithm to detect obstacles and uses an identification method to classify dynamic obstacles.

Although the methods mentioned above have demonstrated success in detecting and tracking dynamic obstacles, their performance is still constrained by inherent sensor limitations. For instance, cameras typically offer a reliable depth range of only up to 5 meters with a limited field of view, while LiDAR's low update frequency, sparse point clouds, and lack of rich features can lead to a high rate of false detections. Inspired by the successful integration of LiDAR and visual sensors in applications such as SLAM \cite{fast-livo} and depth estimation \cite{li2023leovr}, we aim to further enhance dynamic obstacle detection and tracking by leveraging the complementary strengths of both LiDAR and camera sensors.

\section{Methodology} \label{method}
\begin{figure*}[t] 
    \centering
    \includegraphics[scale=0.405]{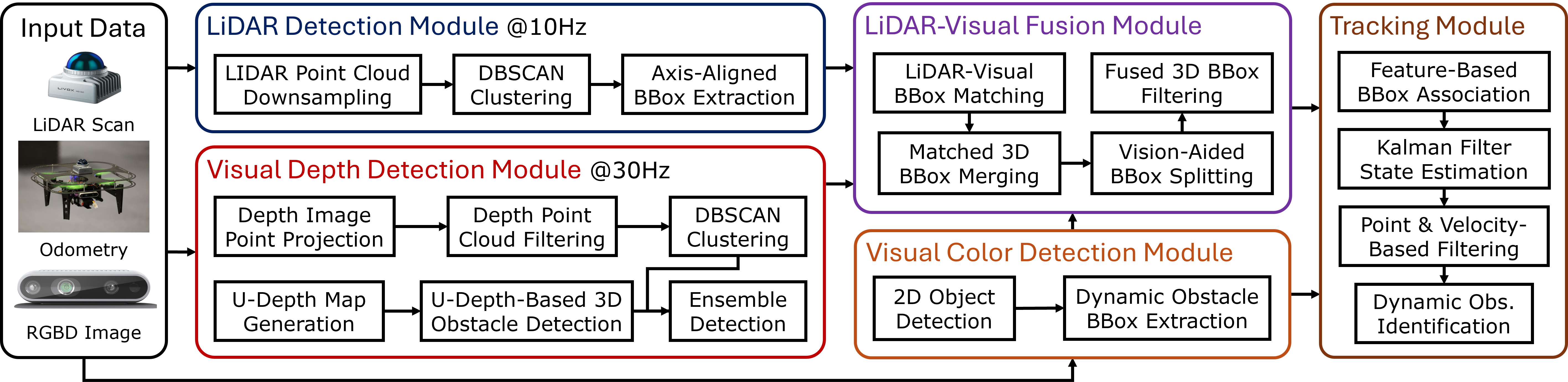}
    \caption{The proposed LiDAR-visual dynamic obstacle detection and tracking framework. Given the LiDAR scan, robot odometry, and RGB-D image, the LiDAR-visual fusion module integrates unclassified obstacle bounding boxes from both the LiDAR and visual (depth and color) detection modules to produce more accurate obstacle detections. The Tracking module then estimates the states of these obstacles and classifies  them as static or dynamic.}
    \label{system_overview}
\end{figure*}

The overall system framework is illustrated in Fig. \ref{system_overview}. Given inputs from the onboard LiDAR, camera, and robot odometry, the LiDAR and visual depth detection modules individually detect unclassified 3D obstacle bounding boxes in the world frame, while the visual color detection module identifies 2D dynamic obstacles in the image plane. These detection results are then fused by the LiDAR-visual Fusion module to generate more accurate obstacle bounding boxes. Finally, the tracking module estimates the states of the detected obstacles and classifies them as static or dynamic.

\subsection{LiDAR Obstacle Detection} \label{LIDAR detection}
\begin{figure}[t] 
    \centering
    \includegraphics[scale=0.49]{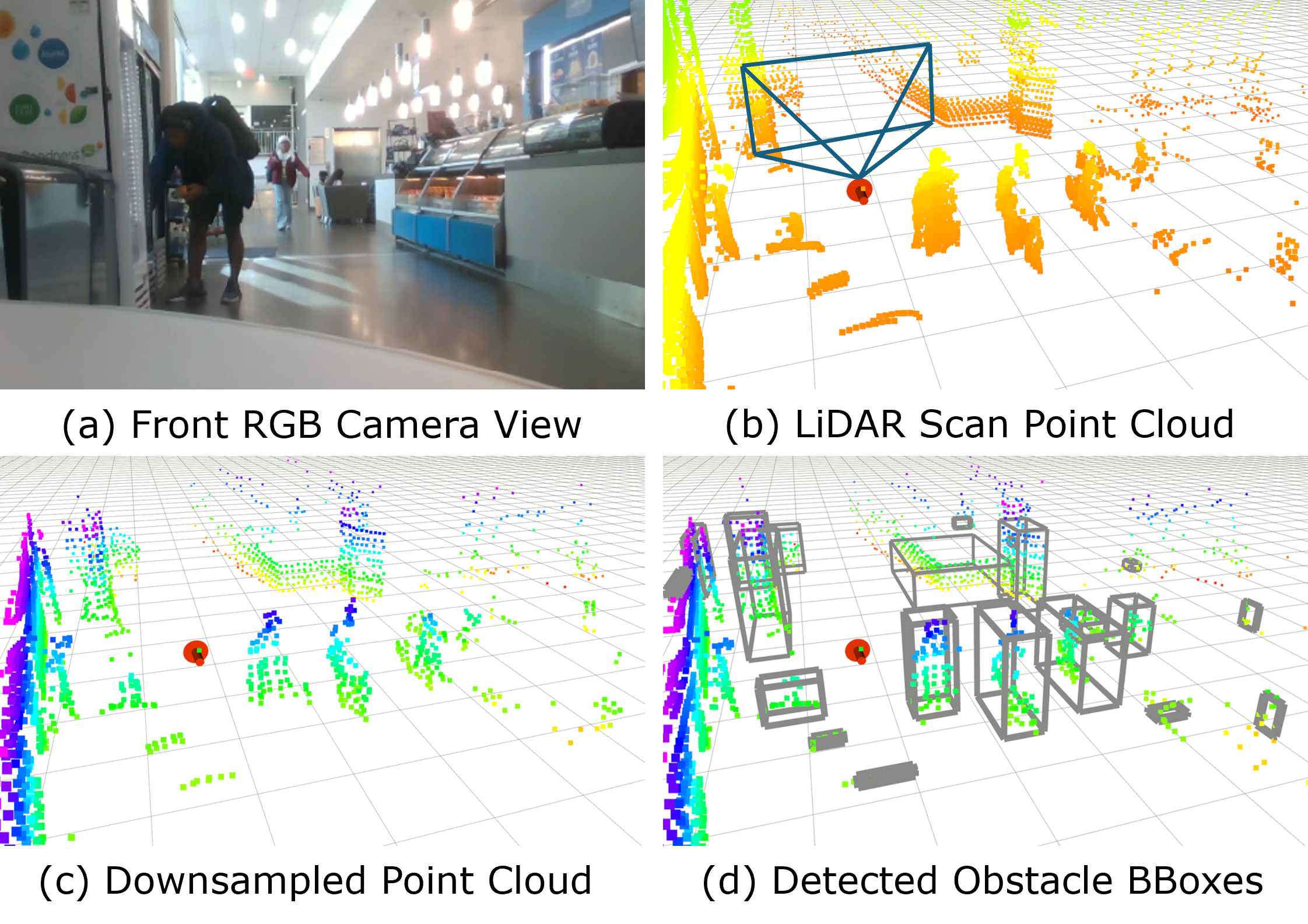}
    \caption{Illustration of the LiDAR obstacle detection process. (a) The scene captured from the front RGB camera. (b) The LiDAR scan point cloud with robot pose and the camera field of view visualized. (c) The downsampled point cloud. (d) The detected obstacle axis-aligned bounding boxes. }
    \label{lidar_illustration}
\end{figure}

The LiDAR detection module extracts 3D axis-aligned bounding boxes for unclassified obstacles from each LiDAR scan. Note that \say{unclassified} means the obstacle has not been identified as static or dynamic by a classification process that is performed by the tracking module as discussed in a later section. The overall process is illustrated in Fig. \ref{lidar_illustration}. Since the original LiDAR scan point cloud shown in Fig. \ref{lidar_illustration}b is dense and low latency is desired, several downsampling filters are applied. First, we use a range filter to retain only points within a specified distance from the robot and then transform the point cloud into the world frame using the robot's odometry. To further reduce the number of points, a distance filter is adopted to retain more points in robot's close proximity while discarding those farther away, based on the following probability formulation for each point $p_{i}$: 
\begin{equation}
    P_{dist}(p_{i}, p_{robot}) = \exp(\frac{-||p_{i} - p_{robot}||_{2}}{\sigma_{dist}^2}),
\end{equation}
where $P_{dist}$ represents the probability of retaining a point, $p_{robot}$ denotes the robot's position, and $\sigma_{dist}$ is a user-specified parameter that controls the distribution of the downsampled points. This distance downsampling filter enables the robot to maintain accurate close-range perception while balancing the computational load. Next, a voxel filter is applied to ensure that the total number of points does not exceed a specified maximum $N_{max}$, which we set to 3000 to balance real-time processing with reasonable accuracy. This results in the downsampled point cloud shown in Fig. \ref{lidar_illustration}c.

With the downsampled point cloud, we adopt the classic density-based clustering algorithm DBSCAN to cluster points, where each cluster corresponds to an obstacle in the environment. To extract axis-aligned 3D bounding boxes, we determine the minimum and maximum values along each axis of the world coordinate and compute the bounding box centers and dimensions accordingly. The resulting unclassified 3D obstacle bounding boxes are shown in Fig. \ref{lidar_illustration}d.

\subsection{Visual Obstacle Detection} \label{visual detection}
The visual obstacle detection comprises both depth and color components, presented as individual modules in Fig. \ref{system_overview}. The visual depth detection module performs a similar function to the LiDAR detection module by generating unclassified 3D obstacle bounding boxes. However, it operates at a higher frequency and achieves greater accuracy for obstacles in close proximity to the robot compared to LiDAR detection. Meanwhile, the visual color detection module leverages the color image to detect 2D bounding boxes for dynamic obstacles. These two components serve different purposes in the subsequent LiDAR-visual fusion module.

\textbf{Visual depth detection module:} Unlike LiDAR detection, which can directly adopt clustering algorithms to obtain unclassified obstacle detections, the inherently noisy depth data from cameras often leads to numerous false positive detections from a similar clustering algorithm. To address this issue, we employ an ensemble detection method \cite{visual_detection} that identifies mutual agreement among the detection results from multiple low-accuracy but computationally efficient detectors. The intuition behind ensemble detection is that while different detectors may produce various errors, detection results common to multiple detectors are likely to be correct. Our visual depth detection module incorporates two detectors: the DBSCAN detector and the U-depth detector, as shown in Fig. \ref{system_overview}. The DBSCAN detector functions similarly to the one used in the LiDAR detection module. In this process, we first project the 2D depth image into a 3D point cloud in the world frame, then apply a voxel filter to reduce the number of points for efficiency, and finally use the DBSCAN clustering algorithm to identify clusters and compute the corresponding 3D obstacle bounding boxes.

\begin{figure}[t] 
    \centering
    \includegraphics[scale=0.73]{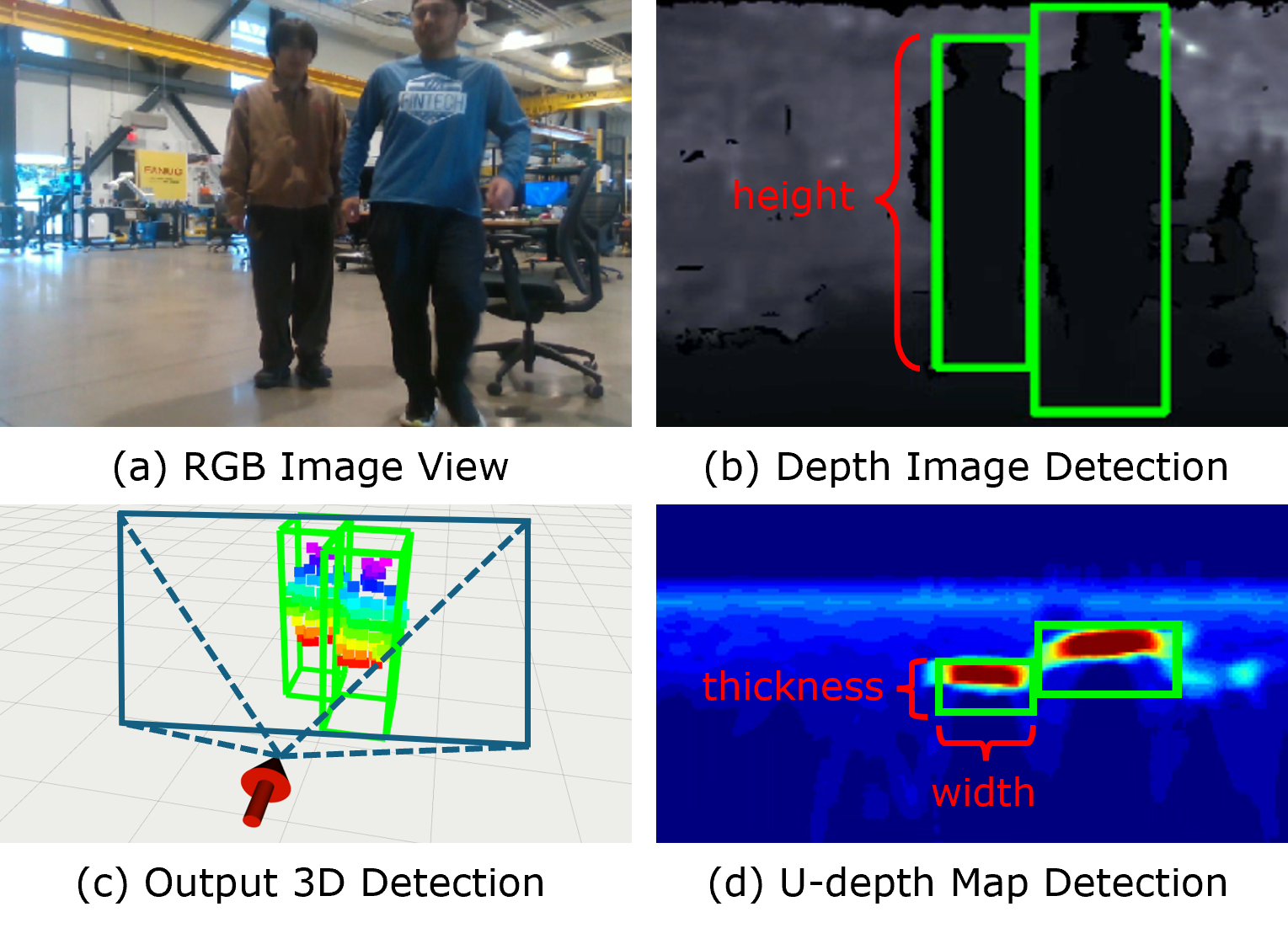}
    \caption{Illustration of the visual U-depth 3D obstacle detection. (a) The RGB image view. (b) The 2D bounding boxes detected in the depth image. (c) The detected 3D bounding boxes with the camera's field of view visualization. (d) The 2D bounding boxes detected from the U-depth map.}
    \label{u_depth_illustration}
\end{figure}

The U-depth detector generates obstacle detection results directly from the depth image, as illustrated in Fig. \ref{u_depth_illustration}. At each time step, a U-depth map (Fig. \ref{u_depth_illustration}d) is produced by computing a histogram of distances from the camera using the depth image, effectively providing a bird's-eye view where the top of the image corresponds to regions closer to the camera. As shown in Fig. \ref{u_depth_illustration}d, regions with obstacles exhibit higher intensity with red color indicated. We apply the line grouping method proposed in \cite{first_udepth} to extract 2D bounding boxes around these regions, which provides the thickness and width of the detected obstacles. With the obstacle widths determined, we compute their heights by analyzing depth continuity, thereby generating 2D bounding boxes on the original depth image (Fig. \ref{u_depth_illustration}b). Finally, by using the 2D bounding box results from the depth image and the U-depth map through coordinate transformation, we obtain the 3D obstacle bounding boxes in the world frame (Fig. \ref{u_depth_illustration}c).

The ensemble detection identifies common detections between the DBSCAN and U-depth detectors by comparing the Intersection over Union (IOU) between their bounding boxes. We retain only the bounding boxes with an IOU exceeding a user-defined threshold and discard the rest as false positives. The matched bounding boxes are then merged by computing the minimal bounding box that encloses both, resulting in the final detected unclassified obstacle bounding box. 

\textbf{Visual color detection module:} In contrast to the LiDAR and visual depth detection modules, which generate unclassified 3D obstacle detections, the visual color detection module directly detects dynamic obstacles by producing 2D bounding boxes in the image. To accomplish this, we apply a pretrained, lightweight, learning-based detection algorithm such as YOLO \cite{yolo} to identify objects in the scene and then extract dynamic obstacles based on predefined classes.

\subsection{LiDAR-visual Detection Fusion}
Using the detection results from the LiDAR and visual modules, the LiDAR-visual detection fusion module combines these outputs to generate refined detection results, as shown in Algorithm \ref{lidar_visual_fusion}. There are three sources of obstacle detections, each offering unique benefits, as described below:
\begin{itemize}
   \item \textit{LiDAR 3D detections} $\mathcal{S}_{\text{lidar}}$: These detections provide the 3D bounding boxes in 360 degrees with long range. 
    \item \textit{Visual depth 3D detections} $\mathcal{S}_{\text{visual3D}}$: These detections offer high-frequency 3D bounding boxes in the robot's close proximity, which is critical for collision avoidance.
    \item \textit{Visual color 2D detections} $\mathcal{S}_{\text{visual2D}}$: These detections can directly yield accurate 2D bounding boxes on the images for dynamic obstacles based on color features.
\end{itemize}

Algorithm \ref{lidar_visual_fusion} begins by reducing false positives within the overlapping detection range of both the LiDAR and camera. In Lines \ref{start_first_for_loop} to \ref{end_of_first_IOU}, we iterate over all visual depth detection results, denoted as \(\mathcal{S}_{\text{visual3D}}\), and match each with the corresponding LiDAR 3D bounding box, \(\mathcal{B}^{i}_{\text{LiDAR}}\), based on the IoU criterion at closely corresponding timestamps. When a match is found (Lines \ref{merge_start}--\ref{merge_end}), we fuse the matched boxes by computing their minimal enclosing bounding box, discarding any visual detections that lack a corresponding LiDAR match as false positives. Since both LiDAR and camera detections involve a clustering step, they can sometimes merge closely spaced obstacles into a single oversized bounding box. To resolve this, we utilize 2D detection results from the color image. In Lines \ref{start_second_for_loop} to \ref{end_of_second_IOU}, we process the fused detections, \(\mathcal{S}_{\text{fused}}\), by computing reprojected 2D bounding boxes, \(\mathcal{B}^{i}_{\text{fused2D}}\), and comparing them with color 2D bounding boxes, \(\mathcal{B}^{i}_{\text{visual2D}}\), using the 2D IoU criterion. If multiple color bounding boxes match the same reprojected 2D bounding box, we split that 2D bounding box into several smaller ones and then project them back into 3D space to generate the final output bounding boxes (Lines \ref{merge_start_2D}--\ref{merge_end_2D}). Finally, we include any remaining LiDAR detections that fall outside the camera’s range in the final outputs. Note that although we assume the robot is typically equipped with one single forward-facing camera, the fusion algorithm is designed to accommodate multiple camera configurations.

\begin{algorithm}[t] \label{lidar_visual_fusion}
\caption{LiDAR-Visual Fusion Algorithm} 
\SetAlgoNoLine%
$\mathcal{S}_{\text{lidar}} \gets \text{LiDAR 3D Obstacle BBoxes}$\;
$\mathcal{S}_{\text{visual3D}}, \mathcal{S}_{\text{visual2D}} \gets \text{Visual Obstacle BBoxes}$\;
$\mathcal{S}_{\text{processed}}, \mathcal{S}_{\text{fused}}, \mathcal{S}_{\text{output}} \gets \{\}$\; 
\For{$\mathcal{B}^{i}_{\normalfont{\text{visual3D}}}$ \normalfont{\textbf{in}} $\mathcal{S}_{\normalfont{\text{visual3D}}}$}{ \label{start_first_for_loop}
    $\mathcal{S}_{\text{match}} \gets \{  \}$\; 
    \For{$\mathcal{B}^{i}_{\normalfont{\text{lidar}}}$ \normalfont{\textbf{in}} $\mathcal{S}_{\normalfont{\text{lidar}}}$}{ 
        $\text{IoU} \gets \normalfont{\textbf{cal3DBBoxIOU}}(\mathcal{B}^{i}_{\normalfont{\text{visual3D}}}, \mathcal{B}^{i}_{\normalfont{\text{lidar}}})$\;
        \If{$\normalfont{\text{IoU}} \geq \normalfont{\text{IoU}_{\text{thresh}}}$}{
            $\mathcal{S}_{\text{match}}.\normalfont{\textbf{append}}(\mathcal{B}^{i}_{\text{lidar}})$\;
            $\mathcal{S}_{\text{processed}}.\normalfont{\textbf{append}}(\mathcal{B}^{i}_{\text{lidar}})$\; \label{end_of_first_IOU}
        }
    }
    \If{$\normalfont{\textbf{len}}(\mathcal{S}_{\normalfont{\text{match}}}) > 1$}{ \label{merge_start}
        $\mathcal{B}_{\text{fused}} \gets \textbf{minEncBBox}(\mathcal{B}^{i}_{\text{visual3D}}, \mathcal{S}_{\text{match}})$\;
        $\mathcal{S}_{\text{fused}}.\textbf{append}(\mathcal{B}_{\text{fused}})$ \label{merge_end}
    }
}

\For{$\mathcal{B}^{i}_{\normalfont{\text{fused}}}$ \normalfont{\textbf{in}} $\mathcal{S}_{\normalfont{\text{fused}}}$}{ \label{start_second_for_loop}
    $\mathcal{B}^{i}_{\text{fused2D}} \gets \textbf{2DReprojection}(\mathcal{B}^{i}_{\text{fused}})$\;
    $\mathcal{S}_{\text{match2D}} \gets \{\}$\;
    \For{$\mathcal{B}^{i}_{\normalfont{\text{visual2D}}}$ \normalfont{\textbf{in}} $\mathcal{S}_{\normalfont{\text{visual2D}}}$}{         
        $\text{IoU}_{\text{2D}} \gets \textbf{cal2DBBoxIOU}(\mathcal{B}^{i}_{\text{visual2D}}, \mathcal{B}^{i}_{\text{fused2D}})$\;
        \If{$\normalfont{\text{IoU}_{\normalfont{\text{2D}}}} \geq \normalfont{\text{IoU}_{\text{thresh2D}}}$}{
            $\mathcal{S}_{\text{match2D}}.\textbf{append}(\mathcal{B}^{i}_{\text{fused2D}})$\; \label{end_of_second_IOU}
        }
    }
    \If{$\normalfont{\textbf{len}}(\mathcal{S}_{\normalfont{\text{match2D}}}) > 1$}{ \label{merge_start_2D}
        $\mathcal{S}_{\text{output}}.\textbf{append}(\textbf{splitBBoxes}(\mathcal{B}^{i}_{\text{fused}}, \mathcal{S}_{\text{match2D}}))$\; 
    }
    \Else{
        $\mathcal{S}_{\text{output}}.\textbf{append}(\mathcal{B}^{i}_{\text{fused}})$\;        \label{merge_end_2D}
    }
}

\For{$\mathcal{B}^{i}_{\normalfont{\text{lidar}}}$ \normalfont{\textbf{in}} $\mathcal{S}_{\normalfont{\text{lidar}}} \ \backslash \  \mathcal{S}_{\normalfont{\text{processed}}}$}{ 
    $\mathcal{S}_{\text{output}}.\textbf{append}(\mathcal{B}^{i}_{\text{lidar}})$\;        
}

\textbf{return} $\mathcal{S}_{\text{output}}$\;
\end{algorithm}

\subsection{Tracking and Dynamic Identification}
The tracking modules run after the LiDAR–visual detection fusion process to further estimate the states of detected obstacles and to identify dynamic obstacles among the unclassified ones. As shown in Fig. \ref{system_overview}, the tracking module begins by performing bounding box association to establish correspondences between obstacles detected at the current time step and those from the previous time step. To achieve this, we construct a feature vector for each detected obstacle, \(O_i\), at both time steps using the following formula:
\begin{equation}
    f(o_{i}) = \begin{bmatrix} \text{pos}(o_{i}) \\ \text{dim}(o_{i}) \\ \text{len}(o_{i}) \\ \text{std}(o_{i}) \end{bmatrix}, \ \text{where} \  f(o_{i}) \in \mathbb{R}^{10} 
\end{equation}
where the feature vector includes the position and dimensions of the detected bounding box as well as the length and standard deviation of the point cloud within that box. For each obstacle detected in the current time step, we compute a similarity score based on the weighted cosine similarity between its feature vector and those of obstacles from the previous time step. If the previous time-step obstacle with the highest similarity also has an IoU with the current obstacle that exceeds a predefined threshold, we consider the match to be valid. Using this feature-based data association prevents the mismatches common with closest center-based methods.

After the association process, we apply a Kalman filter to estimate the states of the detected obstacles, where each state is defined by the obstacle's position, velocity and acceleration. For prediction, the Kalman filter uses a constant acceleration model that captures the changing velocities of dynamic obstacles. The prediction step is formulated as:
\begin{equation}
    \mathbf{X}_{t|t-1} = A \mathbf{X}_{t-1} + Q,
\end{equation}
where $\mathbf{X}$ denotes the obstacle state, \(A\) is the state transition matrix derived from the constant acceleration dynamics, and \(Q\) represents the motion model noise, which is a tunable parameter. For the observation model, we estimate an obstacle's velocity and acceleration from its detected positions:
\[
\mathbf{V}_t = \frac{\mathbf{P}_t - \mathbf{P}_{t-1}}{\delta t}, \quad \mathbf{A}_t = \frac{\mathbf{V}_t - \mathbf{V}_{t-1}}{\delta t},
\]
where \(\mathbf{P}_t\), \(\mathbf{V}_t\), and \(\mathbf{A}_t\) denote the position, velocity, and acceleration of the obstacle, respectively, and \(\delta t\) is the system time step. By combining the prediction and observation models, we apply the standard Kalman filter equations to estimate the final state \(X_t\) for each detected obstacle.

To classify a detected obstacle as dynamic, we apply two complementary checks—if either is satisfied, the obstacle is marked as dynamic. The first check leverages 2D detections from the color image. Since these detections already identify predefined dynamic obstacle classes, we reproject the 3D bounding boxes obtained after Kalman filter estimation onto the 2D image plane and calculate their IoU with the detected 2D bounding boxes. If the IoU exceeds a specified threshold, the obstacle is classified as dynamic. The second check relies on the Kalman filter’s velocity estimations. An obstacle is initially flagged as dynamic if its velocity magnitude exceeds a predefined threshold. However, because errors in position detection can lead to inaccurate velocity estimates and false positives, we supplement this with a point-wise displacement check. For each point within the estimated bounding box, we compute the distance to its closest counterpart in the previous time step’s point cloud. If the majority of these distances exceed a predetermined threshold, the obstacle is classified as dynamic. Otherwise, it is considered static. Combining these two checks results in a more robust identification process.

\section{Result and Discussion}
To evaluate the performance of the proposed dynamic obstacle detection and tracking framework, we conducted both qualitative and quantitative experiments using a collected dataset and physical robot tests. The framework is implemented in ROS and C++. For the physical experiments, we designed a lightweight quadcopter UAV equipped with a forward-facing Intel Realsense D435i RGB-D camera (87° × 58° field of view) and a lightweight Livox Mid-360 LiDAR (360° × 59° field of view). Unlike heavier sensor options, these sensors are inexpensive and lightweight, making them well-suited for mobile robots in a wide range of settings. We use state-of-the-art LiDAR inertial odometry (LIO) \cite{fastlio} for accurate robot localization, and all robot software modules run on the onboard NVIDIA Jetson Orin NX.

In this section, we aim to address the following key questions based on our experimental results:
\begin{itemize}
    \item \textit{How well does the proposed framework perform qualitatively in different environments? Sec. \ref{qualitative_subsection}}. 
    \item \textit{How does the proposed dynamic obstacle detection and tracking algorithm compare to other popular methods in terms of performance? Sec. \ref{benchmark_subsection}}. 
    \item \textit{What benefits do the individual LiDAR and visual modules contribute to the overall framework? Sec. \ref{ablation_subsection}}
    \item \textit{Is this dynamic obstacle perception framework applicable to lightweight mobile robot navigation? Sec. \ref{realworld_subsection}}
\end{itemize}

\subsection{Qualitative Evaluations} \label{qualitative_subsection}
\begin{figure*}[t] 
    \centering
    \includegraphics[scale=0.395]{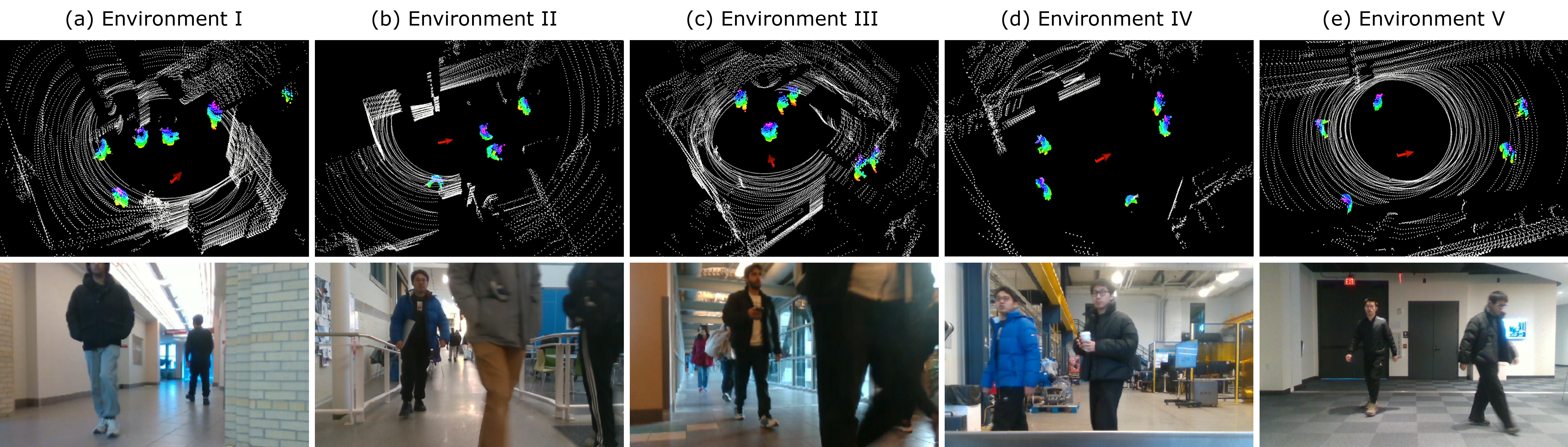}
    \caption{Examples of qualitative experimental results from five testing environments, where the robot, equipped with a LiDAR and a camera, is positioned at the center of each environment. The top row displays point cloud data from the robot’s sensors and highlights detected and tracked dynamic obstacles. The bottom row presents corresponding RGB images from the robot’s camera which provides a visual perspective of the corresponding environments. }
    \label{qualitative_result}
\end{figure*}

To qualitatively evaluate the proposed dynamic obstacle detection and tracking framework, we conduct experiments with the robot in five different environments. Since the mobile robot is primarily used indoors and we are particularly interested in scenarios where it encounters crowded spaces, we select indoor locations with human foot traffic or conduct controlled experiments by having multiple people walk around the robot. The results of dynamic obstacle detection and tracking are visualized in Fig. \ref{qualitative_result}. In these experiments, the top section of Fig. \ref{qualitative_result} displays the point cloud from the robot’s LiDAR and camera, where detected and tracked dynamic obstacles are highlighted. The bottom section presents the corresponding RGB images from the onboard camera. As shown in the figure, all the environments contain approximately five or more dynamic obstacles that pose potential collision risks to the robot, with some also featuring complex static structures. Despite these challenges, the proposed framework successfully detects all dynamic obstacles, demonstrating its adaptability to diverse and dynamic environments. However, we observe occasional misdetections, particularly when obstacles get too close to the LiDAR while remaining outside the camera’s field of view. This issue arises because, at close proximity, the bounding boxes of obstacles become too small, leading to failures in bounding box association. Additionally, when one dynamic obstacle is occluded by another, the occluded obstacle may lose tracking. These limitations can be addressed in future work to enhance the robustness of the framework further.

\subsection{Benchmarking Evaluations} \label{benchmark_subsection}
\begin{figure}[t] 
    \centering
    \includegraphics[scale=0.49]{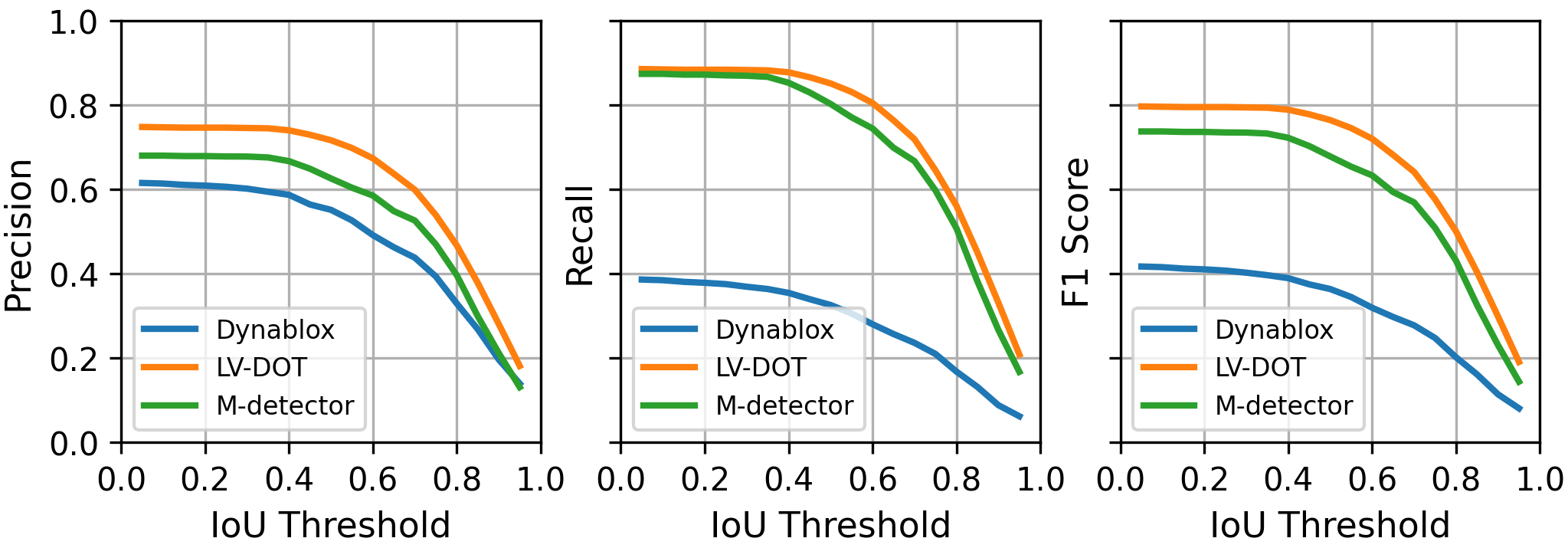}
    \caption{Performance comparison of Dynablox \cite{dynablox}, M-detector \cite{m-detector}, and LV-DOT (Ours) in terms of precision, recall, and F1 score across IoU thresholds. }
    \label{detection_metrics_plot}
\end{figure}

\begin{table*}[t]
\begin{center}
\caption{Benchmarking dynamic obstacle detection performance (precision, recall, F1, and positional error) at three IoU thresholds.} \label{detection_result}
\resizebox{\textwidth}{!}{%
\begin{tabular}{ |c| c c c c| c c c c| c c c c| } 
\hline

 \multirow{2}{*}{Evaluation Metrics} &
  \multicolumn{4}{c|}{$\text{IoU Threshold}=0.3$ (low)} &
  \multicolumn{4}{c|}{$\text{IoU Threshold}=0.5$ (mid)} &
  \multicolumn{4}{c|}{$\text{IoU Threshold}=0.7$ (high)} \Tstrut\\

  & Precision & Recall &  F1  &  Pos. Err.  & Precision & Recall &  F1 &  Pos. Err. & Precision & Recall &  F1  &  Pos. Err.   \Tstrut\\
\hline

 Dynablox \cite{dynablox}  & $60.13\%$  & $36.89\%$ & 0.402 & 0.124m & $55.11\%$  & $32.56\%$ & 0.363 & 0.114m & $43.77\%$  & $23.59\%$ & 0.277 & 0.091m  \Tstrut\\ 
 \hline

 M-detector \cite{m-detector}& $67.74\%$  & $86.89\%$ & 0.734 & 0.094m & $62.56\%$  & $80.21\%$ & 0.678 & 0.085m & $52.58\%$  & $66.67\%$ & 0.569 & 0.071m   \Tstrut\\  
 \hline

 LV-DOT w/o visual &  $79.81\%$  & $80.59\%$ & 0.787 & 0.091m & $74.96\%$  & $75.95\%$ & 0.740 & 0.086m & $60.26\%$  & $61.95\%$ & 0.599 & 0.072m   \Tstrut\\  
 \hline

 LV-DOT w/o LiDAR &  $72.68\%$  & $19.80\%$ & 0.292 & 0.129m & $67.73\%$  & $18.71\%$ & 0.274 & 0.125m & $57.48\%$  & $16.40\%$ & 0.238 & 0.116m   \Tstrut\\  
 \hline
 
 LV-DOT (Ours) &  $74.51\%$  & $88.25\%$ & 0.794 & 0.093m & $71.61\%$  & $85.07\%$ & 0.764 & 0.090m & $59.87\%$  & $71.85\%$ & 0.641 & 0.080m   \Tstrut\\  
 \hline
\end{tabular}%
}
\end{center}
\end{table*}

To quantitatively evaluate the performance of the proposed framework, we conducted benchmarking experiments on our custom dataset. While well-established datasets for 3D object detection in autonomous driving—such as KITTI \cite{kitti} and nuScenes \cite{nuscenes}—exist, they are designed for outdoor applications and do not align with our indoor mobile robotics context. The DOALS \cite{doals} dataset, although suitable in some respects, contains only LiDAR scans and cannot be used to assess our visual detection component. Consequently, we created a new dataset by operating the robot in seven different indoor scenarios that cover a broad range of potential environments, as well as in two motion capture experiments. At a one-second sampling interval, ground truth for dynamic obstacles in the seven indoor scenarios was manually labeled using the SUSTech POINTS \cite{SUSTechPOINTS} labeling tool, and direct ground truth data was obtained from the motion capture environments. Our custom dataset consists of 635 labeled time frames, each containing multiple bounding boxes.

The recent LiDAR-based dynamic object detection algorithms, Dynablox \cite{dynablox} and M-detector \cite{m-detector}, were selected as benchmark methods. For each method, we measured precision, recall, and F1 score for the 3D dynamic obstacle detection bounding boxes at various IoU thresholds and evaluated their average position errors. Detailed evaluation results for IoU thresholds at low (0.3), medium (0.5), and high (0.7) levels are presented in Table \ref{detection_result}, and the precision, recall, and F1 score over IoU thresholds ranging from 0.05 to 0.95 (in 0.05 intervals) are plotted in Fig. \ref{detection_metrics_plot}.

Based on the evaluation results in Table \ref{detection_result}, our method (LV-DOT) outperforms the benchmarks, as evidenced by its consistently highest F1 scores across all IoU thresholds. All methods exhibit similar average position errors of approximately 0.1 meters, and the M-detector achieves the lowest error among the three methods, slightly better than our approach. Comparing our method with the M-detector, both achieve very high and similar recall, indicating that the detection of dynamic obstacles are rarely missed. However, the M-detector's precision is comparatively lower, as our experiments show it often misclassifies static obstacles as dynamic. In contrast, our classification approach, which integrates complementary checks from both visual and LiDAR data, improves overall dynamic obstacle identification accuracy. When comparing our method with Dynablox, we observe that Dynablox exhibits lower precision and recall. This is primarily because Dynablox, designed for denser LiDAR data, struggles with the sparse point clouds produced by lightweight LiDAR sensors, which limits its ability to capture object motion. In contrast, our method is better suited for lightweight sensors and mobile robots. Similar results are observed in Fig. \ref{detection_metrics_plot}, where our method achieves overall higher F1 scores compared to the other two methods, maintaining high precision and recall across all IoU thresholds.

\subsection{Ablation Study} \label{ablation_subsection}
Since the proposed framework integrates both LiDAR and visual components, we analyze the contribution of each to the overall performance. Table \ref{detection_result} presents benchmarking experiments for the complete framework, as well as for variants without the visual module (LV-DOT w/o visual) and without the LiDAR module (LV-DOT w/o LiDAR). Comparing the full framework with the variant that excludes the visual module, we observe an overall improvement, indicated by higher F1 scores across all IoU thresholds. Specifically, the visual component significantly enhances recall, which is expected because our RGB image detection aids dynamic obstacle classification, and the camera's denser point cloud in close proximity helps reduce missed detections. However, we notice a slight drop in precision and average position accuracy, which is likely due to the camera's less accurate depth estimation compared to LiDAR. Additionally, the visual module operates at a higher frequency, providing a faster response to dynamic obstacles during navigation. Besides, when comparing the framework with and without the LiDAR component, both precision and recall are markedly improved and average position errors are reduced. This improvement is attributable to LiDAR's higher accuracy in distance estimation and its ability to detect obstacles from all directions.

\subsection{Physical Robot Experiments} \label{realworld_subsection}

\begin{table}[t]
    \centering
    \caption{Runtime of Each Module in the Proposed System}
    \begin{tabular}{lcc}
    \hline
    \textbf{System Modules} & \textbf{Mean (ms)} & \textbf{Std. Dev. (ms)} \Tstrut\\
    \hline
    LiDAR Detection           & 8.34  & 4.15 \Tstrut\\ 
    Visual Depth Detection    & 10.22  & 3.46 \Tstrut\\ 
    Visual Color Detection    & 34.15 & 0.91 \Tstrut\\ 
    LiDAR-Visual Fusion       & 0.40 & 0.13 \Tstrut\\ 
    Tracking and Identification & 2.29 & 0.75 \Tstrut\\ 
    \hline
    \end{tabular}%
    \label{runtime_table}
\end{table}

\begin{figure}[t] 
    \centering
    \includegraphics[scale=0.36]{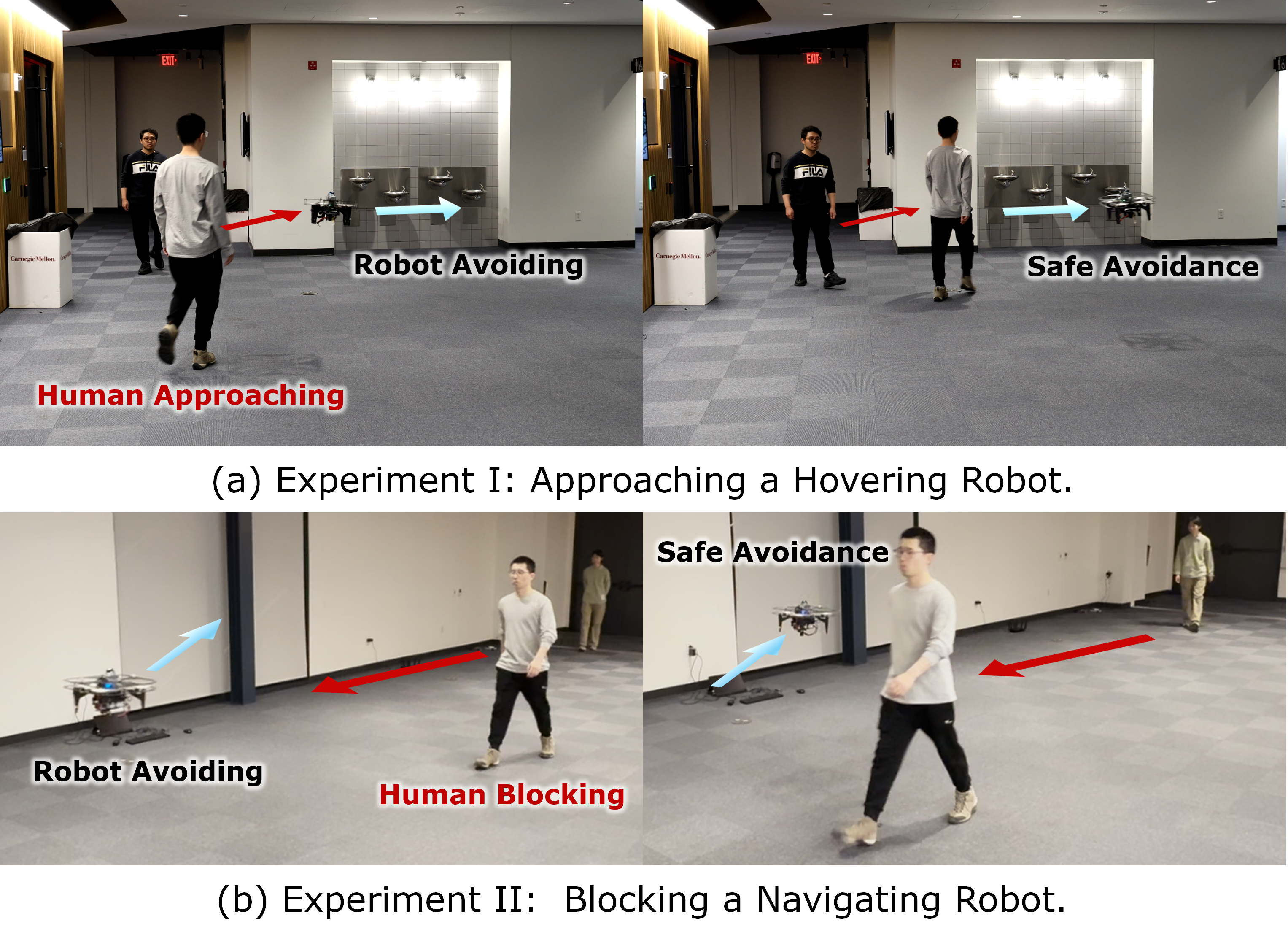}
    \caption{Illustration of the physical robot experiments. We conduct two types of experiments to demonstrate the applicability of the proposed framework in robot navigation and collision avoidance: (a) A human approaches a hovering robot. (b) A human blocks the path of a navigating robot. }
    \label{real_experiment}
\end{figure}

To verify the effectiveness of the proposed framework in real-world robot navigation applications, we conduct physical robot experiments, as illustrated in Fig. \ref{real_experiment}. The experiments consist of two scenarios, utilizing the planner from \cite{navrl} for navigation and collision avoidance. In the first experiment, a quadcopter hovers in place while several humans deliberately walk toward it, attempting to collide with the robot. Fig. \ref{real_experiment}a captures a moment when a human approaches the robot, after which the robot successfully dodges and moves to a safe position. In the second experiment, the robot follows a predefined path to simulate a navigation scenario while humans intentionally walk to block its path. Fig. \ref{real_experiment}b illustrates a successful example where the robot avoids collisions while effectively approaching its goal. Both experiments demonstrate that the proposed framework enables real-time detection and tracking of dynamic obstacles, allowing the robot to navigate safely and avoid collisions.

During the physical robot experiments and dataset evaluation, we recorded the runtime of the proposed system on the robot's onboard computer, as shown in Table \ref{runtime_table}. The table presents the mean runtime and standard deviation for each module. The LiDAR detection and visual depth detection modules operate in approximately 8.34 ms and 10.22 ms, respectively. The visual color detection module, which requires more computational resources, runs at an average of 34.15 ms, running at a rate close to 30 Hz. The LiDAR-visual fusion and tracking modules are highly efficient, with mean runtimes of 0.40 ms and 2.29 ms, respectively. Since these modules can be executed in parallel, the overall system achieves real-time performance on actual robots, meeting the critical requirement for dynamic obstacle perception and navigation in real-world environments.

\section{Conclusion and Future Work} 
This paper presents a LiDAR-visual dynamic obstacle detection and tracking framework for robot navigation that addresses the limitations of single-sensor systems. The method begins by independently detecting obstacles from LiDAR scans, depth images, and color images, using specialized engineering techniques to achieve accurate 3D detection of unclassified obstacles from LiDAR and depth data. To fully leverage the strengths of each sensor, we introduce a LiDAR-visual fusion algorithm that refines detection results by integrating data from all sources. Additionally, a tracking module comprising feature-based data association and Kalman filtering generates the final set of dynamic obstacles using our dynamic identification method. The framework has been thoroughly evaluated both qualitatively and quantitatively, demonstrating superior performance compared to single-sensor approaches on the dataset. We also integrated the framework into an autonomous robot and verified its effectiveness for navigation. 
Although the proposed framework demonstrates satisfactory real-time detection and tracking of dynamic obstacles, there remains room for improvement in reducing false positives and negatives, which we plan to address in future work to improve reliability further.

\bibliographystyle{IEEEtran}
\bibliography{bibliography.bib}

\begin{thebibliography}{10}
\providecommand{\url}[1]{#1}
\csname url@samestyle\endcsname
\providecommand{\newblock}{\relax}
\providecommand{\bibinfo}[2]{#2}
\providecommand{\BIBentrySTDinterwordspacing}{\spaceskip=0pt\relax}
\providecommand{\BIBentryALTinterwordstretchfactor}{4}
\providecommand{\BIBentryALTinterwordspacing}{\spaceskip=\fontdimen2\font plus
\BIBentryALTinterwordstretchfactor\fontdimen3\font minus \fontdimen4\font\relax}
\providecommand{\BIBforeignlanguage}[2]{{%
\expandafter\ifx\csname l@#1\endcsname\relax
\typeout{** WARNING: IEEEtran.bst: No hyphenation pattern has been}%
\typeout{** loaded for the language `#1'. Using the pattern for}%
\typeout{** the default language instead.}%
\else
\language=\csname l@#1\endcsname
\fi
#2}}
\providecommand{\BIBdecl}{\relax}
\BIBdecl

\bibitem{second}
Y.~Yan, Y.~Mao, and B.~Li, ``Second: Sparsely embedded convolutional detection,'' \emph{Sensors}, vol.~18, no.~10, p. 3337, 2018.

\bibitem{pv-rcnn}
S.~Shi, C.~Guo, L.~Jiang, Z.~Wang, J.~Shi, X.~Wang, and H.~Li, ``Pv-rcnn: Point-voxel feature set abstraction for 3d object detection,'' in \emph{Proceedings of the IEEE/CVF conference on computer vision and pattern recognition}, 2020, pp. 10\,529--10\,538.

\bibitem{vision_ccmpc}
J.~Lin, H.~Zhu, and J.~Alonso-Mora, ``Robust vision-based obstacle avoidance for micro aerial vehicles in dynamic environments,'' in \emph{2020 IEEE International Conference on Robotics and Automation (ICRA)}.\hskip 1em plus 0.5em minus 0.4em\relax IEEE, 2020, pp. 2682--2688.

\bibitem{zju_dynamic_avoidance}
Y.~Wang, J.~Ji, Q.~Wang, C.~Xu, and F.~Gao, ``Autonomous flights in dynamic environments with onboard vision,'' in \emph{2021 IEEE/RSJ International Conference on Intelligent Robots and Systems (IROS)}.\hskip 1em plus 0.5em minus 0.4em\relax IEEE, 2021, pp. 1966--1973.

\bibitem{dsp}
G.~Chen, W.~Dong, P.~Peng, J.~Alonso-Mora, and X.~Zhu, ``Continuous occupancy mapping in dynamic environments using particles,'' \emph{IEEE Transactions on Robotics}, 2023.

\bibitem{m-detector}
H.~Wu, Y.~Li, W.~Xu, F.~Kong, and F.~Zhang, ``Moving event detection from lidar point streams,'' \emph{nature communications}, vol.~15, no.~1, p. 345, 2024.

\bibitem{dynablox}
L.~Schmid, O.~Andersson, A.~Sulser, P.~Pfreundschuh, and R.~Siegwart, ``Dynablox: Real-time detection of diverse dynamic objects in complex environments,'' \emph{IEEE Robotics and Automation Letters}, 2023.

\bibitem{visual_detection}
Z.~Xu, X.~Zhan, Y.~Xiu, C.~Suzuki, and K.~Shimada, ``Onboard dynamic-object detection and tracking for autonomous robot navigation with rgb-d camera,'' \emph{IEEE Robotics and Automation Letters}, vol.~9, no.~1, pp. 651--658, 2024.

\bibitem{dot}
I.~Ballester, A.~Font{\'a}n, J.~Civera, K.~H. Strobl, and R.~Triebel, ``Dot: Dynamic object tracking for visual slam,'' in \emph{2021 IEEE international conference on robotics and automation (ICRA)}.\hskip 1em plus 0.5em minus 0.4em\relax IEEE, 2021, pp. 11\,705--11\,711.

\bibitem{ea-lss}
H.~Hu, F.~Wang, J.~Su, Y.~Wang, L.~Hu, W.~Fang, J.~Xu, and Z.~Zhang, ``Ea-lss: Edge-aware lift-splat-shot framework for 3d bev object detection,'' \emph{arXiv preprint arXiv:2303.17895}, 2023.

\bibitem{eth_yolo_detection}
T.~Eppenberger, G.~Cesari, M.~Dymczyk, R.~Siegwart, and R.~Dub{\'e}, ``Leveraging stereo-camera data for real-time dynamic obstacle detection and tracking,'' in \emph{2020 IEEE/RSJ International Conference on Intelligent Robots and Systems (IROS)}.\hskip 1em plus 0.5em minus 0.4em\relax IEEE, 2020, pp. 10\,528--10\,535.

\bibitem{first_udepth}
H.~Oleynikova, D.~Honegger, and M.~Pollefeys, ``Reactive avoidance using embedded stereo vision for mav flight,'' in \emph{2015 IEEE International Conference on Robotics and Automation (ICRA)}.\hskip 1em plus 0.5em minus 0.4em\relax IEEE, 2015, pp. 50--56.

\bibitem{dynamic_map_ours}
Z.~Xu, X.~Zhan, B.~Chen, Y.~Xiu, C.~Yang, and K.~Shimada, ``A real-time dynamic obstacle tracking and mapping system for uav navigation and collision avoidance with an rgb-d camera,'' in \emph{2023 IEEE International Conference on Robotics and Automation (ICRA)}, 2023, pp. 10\,645--10\,651.

\bibitem{saha2022efficient}
A.~Saha, B.~C. Dhara, S.~Umer, K.~Yurii, J.~M. Alanazi, and A.~A. AlZubi, ``Efficient obstacle detection and tracking using rgb-d sensor data in dynamic environments for robotic applications,'' \emph{Sensors}, vol.~22, no.~17, p. 6537, 2022.

\bibitem{lu2022perception}
M.~Lu, H.~Chen, and P.~Lu, ``Perception and avoidance of multiple small fast moving objects for quadrotors with only low-cost rgbd camera,'' \emph{IEEE Robotics and Automation Letters}, vol.~7, no.~4, pp. 11\,657--11\,664, 2022.

\bibitem{chen2022real}
H.~Chen and P.~Lu, ``Real-time identification and avoidance of simultaneous static and dynamic obstacles on point cloud for uavs navigation,'' \emph{Robotics and Autonomous Systems}, vol. 154, p. 104124, 2022.

\bibitem{sun2017improving}
Y.~Sun, M.~Liu, and M.~Q.-H. Meng, ``Improving rgb-d slam in dynamic environments: A motion removal approach,'' \emph{Robotics and Autonomous Systems}, vol.~89, pp. 110--122, 2017.

\bibitem{shi2018dynamic}
W.~Shi, J.~Li, Y.~Liu, D.~Zhu, D.~Yang, and X.~Zhang, ``Dynamic obstacles rejection for 3d map simultaneous updating,'' \emph{IEEE Access}, vol.~6, pp. 37\,715--37\,724, 2018.

\bibitem{ds_slam}
C.~Yu, Z.~Liu, X.-J. Liu, F.~Xie, Y.~Yang, Q.~Wei, and Q.~Fei, ``Ds-slam: A semantic visual slam towards dynamic environments,'' in \emph{2018 IEEE/RSJ international conference on intelligent robots and systems (IROS)}.\hskip 1em plus 0.5em minus 0.4em\relax IEEE, 2018, pp. 1168--1174.

\bibitem{dai2020rgb}
W.~Dai, Y.~Zhang, P.~Li, Z.~Fang, and S.~Scherer, ``Rgb-d slam in dynamic environments using point correlations,'' \emph{IEEE transactions on pattern analysis and machine intelligence}, vol.~44, no.~1, pp. 373--389, 2020.

\bibitem{asvadi20163d}
A.~Asvadi, C.~Premebida, P.~Peixoto, and U.~Nunes, ``3d lidar-based static and moving obstacle detection in driving environments: An approach based on voxels and multi-region ground planes,'' \emph{Robotics and Autonomous Systems}, vol.~83, pp. 299--311, 2016.

\bibitem{yoon2019mapless}
D.~Yoon, T.~Tang, and T.~Barfoot, ``Mapless online detection of dynamic objects in 3d lidar,'' in \emph{2019 16th Conference on Computer and Robot Vision (CRV)}.\hskip 1em plus 0.5em minus 0.4em\relax IEEE, 2019, pp. 113--120.

\bibitem{dewan2016motion}
A.~Dewan, T.~Caselitz, G.~D. Tipaldi, and W.~Burgard, ``Motion-based detection and tracking in 3d lidar scans,'' in \emph{2016 IEEE international conference on robotics and automation (ICRA)}.\hskip 1em plus 0.5em minus 0.4em\relax IEEE, 2016, pp. 4508--4513.

\bibitem{min2021kernel}
Y.~Min, D.-U. Kim, and H.-L. Choi, ``Kernel-based 3-d dynamic occupancy mapping with particle tracking,'' in \emph{2021 IEEE International Conference on Robotics and Automation (ICRA)}.\hskip 1em plus 0.5em minus 0.4em\relax IEEE, 2021, pp. 5268--5274.

\bibitem{zhou2023lidar}
Z.~Zhou, X.~Feng, S.~Di, and X.~Zhou, ``A lidar mapping system for robot navigation in dynamic environments,'' \emph{IEEE Transactions on Intelligent Vehicles}, 2023.

\bibitem{saha20243d}
A.~Saha and B.~C. Dhara, ``3d lidar-based obstacle detection and tracking for autonomous navigation in dynamic environments,'' \emph{International Journal of Intelligent Robotics and Applications}, vol.~8, no.~1, pp. 39--60, 2024.

\bibitem{du2024lidar}
W.~Du and G.~Beltrame, ``Lidar-based real-time object detection and tracking in dynamic environments,'' \emph{arXiv preprint arXiv:2407.04115}, 2024.

\bibitem{fast-livo}
C.~Zheng, W.~Xu, Z.~Zou, T.~Hua, C.~Yuan, D.~He, B.~Zhou, Z.~Liu, J.~Lin, F.~Zhu \emph{et~al.}, ``Fast-livo2: Fast, direct lidar-inertial-visual odometry,'' \emph{IEEE Transactions on Robotics}, 2024.

\bibitem{li2023leovr}
D.~Li, J.~Xu, Z.~Yang, Q.~Ma, L.~Zhang, and P.~Chen, ``Leovr: Motion-inspired visual-lidar fusion for environment depth estimation,'' \emph{IEEE Transactions on Mobile Computing}, 2023.

\bibitem{yolo}
J.~Redmon, S.~Divvala, R.~Girshick, and A.~Farhadi, ``You only look once: Unified, real-time object detection,'' in \emph{2016 IEEE Conference on Computer Vision and Pattern Recognition (CVPR)}, 2016, pp. 779--788.

\bibitem{fastlio}
W.~Xu, Y.~Cai, D.~He, J.~Lin, and F.~Zhang, ``Fast-lio2: Fast direct lidar-inertial odometry,'' \emph{IEEE Transactions on Robotics}, vol.~38, no.~4, pp. 2053--2073, 2022.

\bibitem{kitti}
A.~Geiger, P.~Lenz, C.~Stiller, and R.~Urtasun, ``Vision meets robotics: The kitti dataset,'' \emph{The International Journal of Robotics Research}, vol.~32, no.~11, pp. 1231--1237, 2013.

\bibitem{nuscenes}
H.~Caesar, V.~Bankiti, A.~H. Lang, S.~Vora, V.~E. Liong, Q.~Xu, A.~Krishnan, Y.~Pan, G.~Baldan, and O.~Beijbom, ``nuscenes: A multimodal dataset for autonomous driving,'' in \emph{Proceedings of the IEEE/CVF conference on computer vision and pattern recognition}, 2020, pp. 11\,621--11\,631.

\bibitem{doals}
P.~Pfreundschuh, H.~F. Hendrikx, V.~Reijgwart, R.~Dub{\'e}, R.~Siegwart, and A.~Cramariuc, ``Dynamic object aware lidar slam based on automatic generation of training data,'' in \emph{2021 IEEE International Conference on Robotics and Automation (ICRA)}.\hskip 1em plus 0.5em minus 0.4em\relax IEEE, 2021, pp. 11\,641--11\,647.

\bibitem{SUSTechPOINTS}
E.~Li, S.~Wang, C.~Li, D.~Li, X.~Wu, and Q.~Hao, ``Sustech points: A portable 3d point cloud interactive annotation platform system,'' in \emph{2020 IEEE Intelligent Vehicles Symposium (IV)}, 2020, pp. 1108--1115.

\bibitem{navrl}
Z.~Xu, X.~Han, H.~Shen, H.~Jin, and K.~Shimada, ``Navrl: Learning safe flight in dynamic environments,'' \emph{arXiv preprint arXiv:2409.15634}, 2024.

\end{thebibliography}

\end{document}